\def\year{2019}\relax
\documentclass[letterpaper]{article} % DO NOT CHANGE THIS
\usepackage{aaai19}  % DO NOT CHANGE THIS
\usepackage{times}  % DO NOT CHANGE THIS
\usepackage{helvet} % DO NOT CHANGE THIS
\usepackage{courier}  % DO NOT CHANGE THIS
\usepackage[hyphens]{url}  % DO NOT CHANGE THIS
\usepackage{graphicx} % DO NOT CHANGE THIS
\usepackage{graphicx}  % DO NOT CHANGE THIS
\title{Where is My Stuff? An Interactive System for Spatial Relations }
\author{ \Large \textbf{E. Akin Sisbot, Jonathan H. Connell }
\\ % All authors must be in the same font size and format. Use \Large and \textbf to achieve this result when breaking a line
IBM Research\\ %If you have multiple authors and multiple affiliations
IBM T.J. Watson Research Center\\
Yorktown Heights, NY 10598\\
\{easisbot, jconnell\}@us.ibm.com % email address must be in roman text type, not monospace or sans serif
}
 
\begin{document}

\maketitle

\begin{abstract}

In this paper we present a system that detects and tracks objects and agents, computes spatial relations, and communicates those relations to the user using speech. Our system is able to detect multiple objects and agents at 30 frames per second using a RGBD camera. It is able to extract the spatial relations \textit{in, on, next to, near}, and \textit{belongs to}, and communicate these relations using natural language. The notion of belonging is particularly important for Human-Robot Interaction since it allows the robot ground the language and reason about the right objects.
Although our system is currently static and targeted to a fixed location in a room, we are planning to port it to a mobile robot thus allowing it explore the environment and create a spatial knowledge base.

%Understanding the spatial relations between objects and translating them into natural language is a key component of a robot to have to solve

\end{abstract}

\section{Introduction}

With the recent developments in Artificial Intelligence and adoption of smart devices, social robots are one step closer to finding a place in our daily lives. From doing repetitive mundane tasks to helping us to solve problems, they will be a major contributor to improving our quality of life. One problem that social robots/agents can help with is to keeping track of objects that we use in our homes or workplaces.  Forgetting where an object is, or somebody else moving an object without our knowledge, are two very common scenarios that are sources of frustration. We believe that an interactive camera system along with adequate reasoning capabilities can be very helpful in solving this problem. 

In this paper we present an integrated system that constantly watches the scene, detecting and tracking objects and people while inferring ownership and spatial relationships. The system allows the user ask where an object is, and answers accordingly. Although the system is currently installed as a static setup as part of a smart room, our vision is to port it to a mobile robot.

Spatial relationships between objects and people have been studied extensively in computer vision and developmental psychology fields. Piaget \cite{Piaget54} found that the notion of spatial relationships between objects starts in early infanthood. This ability is crucial for us to build an abstract representation of the world and communicate this representation with others.

In robotics, spatial representations are related to affordances. In \cite{Rosman11} a contact point network is used to segment out objects using a Kinect camera, and the relationships are learned using supervised learning methods. \cite{Fichtl13} use a 3D simulation environment to generate large volume of training data. Some works such as \cite{Belz15} use 2D images instead of 3D to calculate spatial relationships.

\cite{Mees17} described a system where the robot is able to use its previous knowledge to create new relationships between objects. In a recent work by \cite{Jund18}, a neural network is employed to generalize spatial relations to apply to previously unknown objects. Recent VQA systems such as \cite{cho17} employ deep neural networks to answer any questions about a scene.

In this paper we present an interactive system that uses a simplified approach to calculate spatial relations. Our system is able to calculate belonging relationship simplifying the  language that the user use to refer to objects. However our system offers an end-to-end solution starting from user speech and ending with system's answer in natural language.

\section {Use Cases}

There are two use cases that inspire and guide our work:

\subsection{Elder Care}

About 40\% of people aged 65 or older have age-associated memory impairment \cite{small02}. The older we get the easier it becomes to forget where we placed our belongings. This use case focuses on keeping track of relevant items that an elderly person might need but forgot. 
An example scenario for this use case is as follows:

\begin{itemize}
\item Mr. Jones comes home, leaves his wallet in an uncommon location (next to the vase)
\item Ms. Jones places a couple of magazines on the wallet
\item Later Mr. Jones asks ``Where is my wallet?"
\item The system answers: ``It is next to the vase, under the magazines"
\end{itemize}

\subsection{Workshops and Factories}

In workplaces, factories, and workshops a missing or misplaced object may have a direct impact on the efficiency of the processes underway. In this use case the system not only answers explicit queries but also proactively watches and warns the user of misplaced/missing items. 
An example for this use case might be:

\begin{itemize}
\item Mr. Jones is working on a repair project
\item The current project involves a defined set of tools
\item The system tracks the location of those tools
\item The system detects that one tool is missing from its usual location
\begin{itemize}
\item The system warns Mr. Jones that the tool is missing
\end{itemize}
\item The system detects that one tool is situated in an unusual location
\begin{itemize}
\item When the time comes for Mr. Jones to use that tool, the system proactively tells him its location relative to a landmark: ``The wrench is behind the toolbox"
\end{itemize}
\end{itemize}

\section{System Overview}

\begin{figure}[!ht]
  \centering
  \includegraphics[width=0.8\columnwidth]{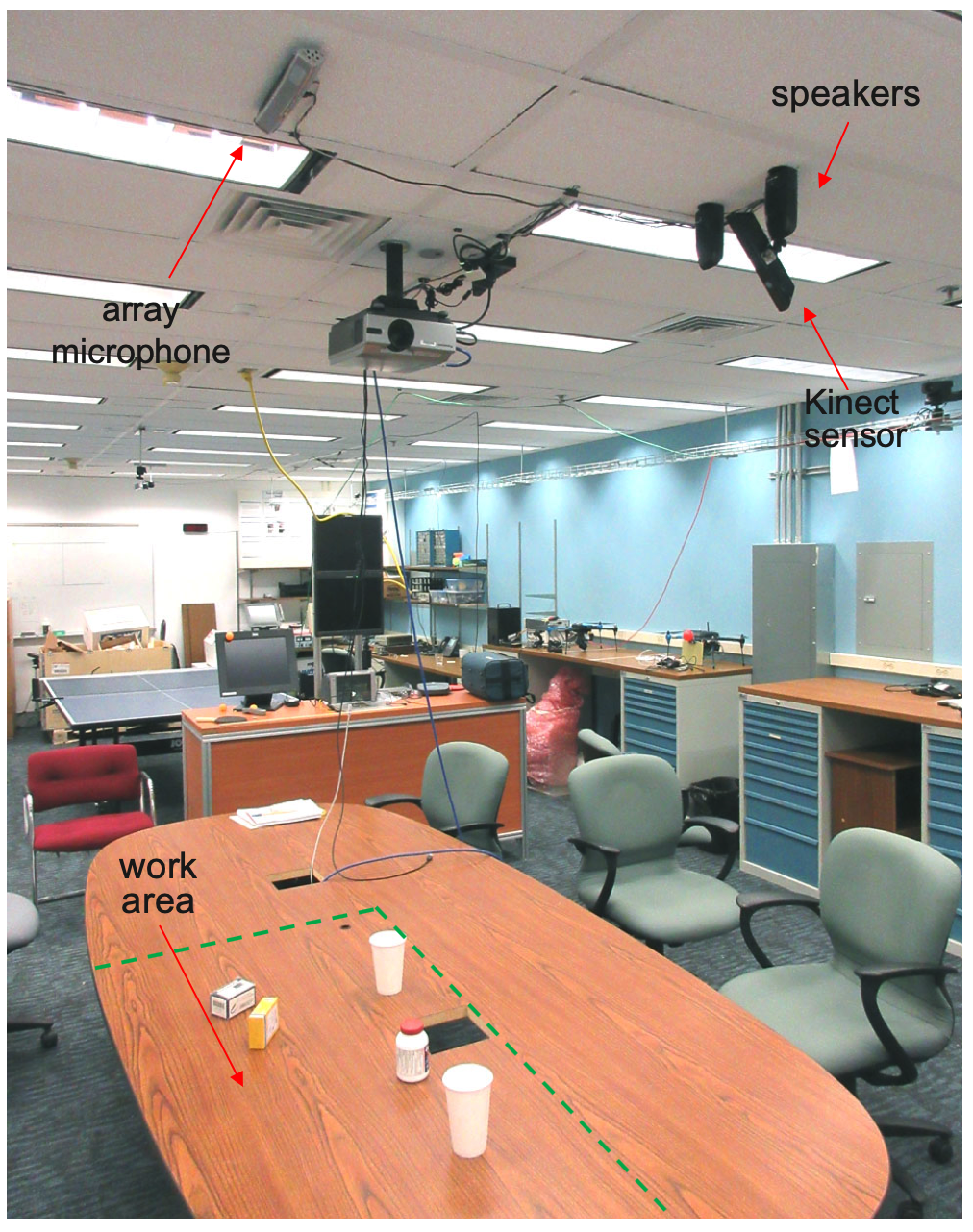}
  \caption{The system uses a MS Kinect sensor and an array microphone to detect objects and capture speech.}
  \label{fig:setup}
\end{figure}

Our system uses a Microsoft Kinect RGB-D camera and an array microphone mounted on the ceiling. As seen in Figure \ref{fig:setup}, the camera is focused on a target work area where we detect and track the objects. The work area can easily be extended by using multiple cameras.

\noindent There are three major components of the system:

\subsection{Object and Person Detection}

People are essentially modeled as stalagmites coming up from the floor, with constraints on head height, head size, shoulder width, etc. Objects are similarly modeled as bumps coming up from the work surface. Both people and objects are tracked over time allowing continuity in properties such as names or types that may have been assigned. The full set of person and object data is streamed in JSON format via a ZeroMQ pub-sub channel approximately 30 times a second. 

The system allows objects to be tracked even when they are being held (and thus have a wildly different shape when combined with an arm) by merging the point cloud associated with the user's hand with the point cloud of the object.  
The system also tracks the position of people's hands relative to the work area focusing on the end points of the protrusions in the point cloud that are created by the presence of arms). 
In addition to providing information about hands, the system can infer pointing directions using principal component analysis method. This can be used by itself to, for instance, select a particular object for querying. 

\begin{figure}[!ht]
  \centering
  \includegraphics[width=0.8\columnwidth]{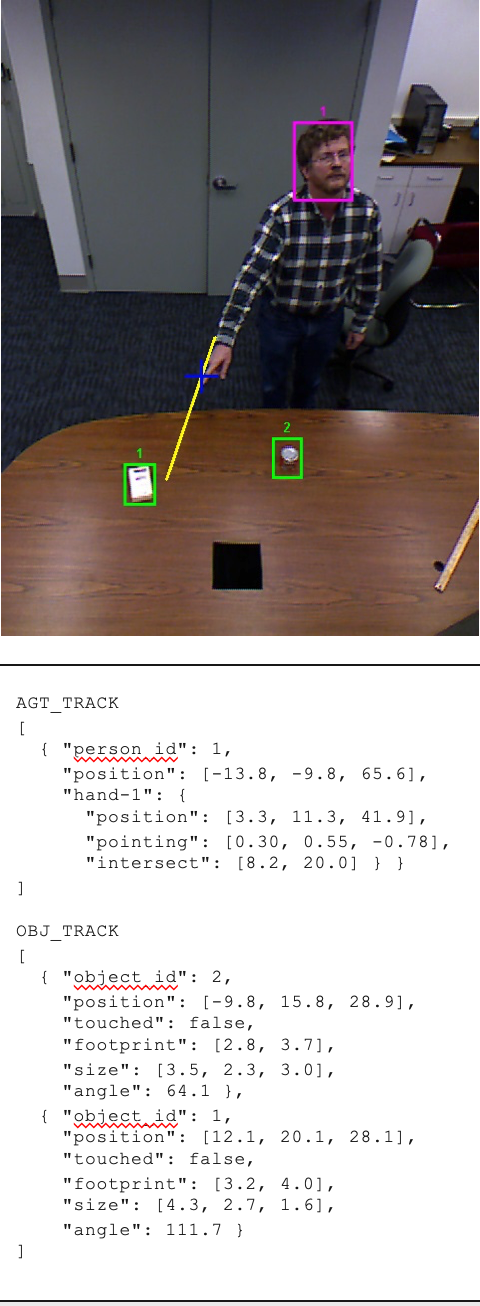}
  \caption{A snapshot of the environment and the ZeroMQ detection stream data.}
  \label{fig:zmq}
\end{figure}

Figure \ref{fig:zmq} shows a snapshot of the environment and the corresponding person and object detection results.  The results are then passed to the Spatial Relations step.

\subsection{Spatial Relations}

In order to generate human understandable dialog, we extract the topology of the environment by calculating relationships between agents, objects, and locations from the geometric data. For every detection result that we receive we compute a number of spatial relationships.

We consider 4 types of observer-independent spatial relations:
\begin{itemize}
\item Object-Object Relations: in, on, near, next to
\item Object-Agent Relations: belongs, last touched by
\item Agent-Location Relations: in
\item Object-Location Relations: in
\end{itemize}

\begin{figure}[!ht]
  \centering
  \includegraphics[width=\columnwidth]{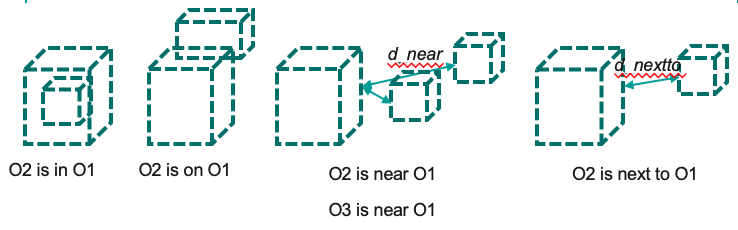}
  \caption{Illustration of Object to Object Relations}
  \label{fig:rel}
\end{figure}

These spatial relations are computed based on a set of predefined rules and a number of geometric properties. Figure~\ref{fig:rel} illustrates Object-Object relations. 

\subsubsection{In} The \textit{in} relation between object $O_1$ and $O_2$ is defined as:

$O_1$ is \textit{in} $O_2$ if 80\%  of $O_1$'s volume is in $O_2$.

\subsubsection{On} The \textit{on} relation between object $O_1$ and $O_2$ is defined as:

$O_1$ is \textit{on} $O_2$ if the bottom of $O_1$ is above the top of $O_2$.

\subsubsection{Near} The \textit{near} relation between object $O_1$ and $O_2$ is defined as:
$O_1$ is \textit{near} $O_2$ if the distance between $O_1$ and $O_2$ is not greater than the $max(x_{O_1},y_{O_1},z_{O_1},x_{O_2},y_{O_2},z_{O_2}) $.

\subsubsection{Next to} The \textit{next to} relation between object $O_1$ and $O_2$ is defined as:

$O_1$ is \textit{next to} $O_1$ if $O_1$ is \textit{near} $O_1$ and there no object between $O_1$ and $O_2$.

\subsubsection{Belongs} The object $O_1$ belongs to the agent $A_1$ if $O_1$ has not been seen before and appears with the agent $A_1$

The spatial relations are computed for every frame and for every object and agent. The system is also able to compute only a subset of these relations to focus only on relevant objects and agents, if desired. The relationships, especially \textit{belongs to}, is kept and maintained in a database allowing the user ask about his/her/others' objects.

\subsection{Dialog}

The Dialog system is the third major component of the overall architecture. It is in charge of understanding the user's spoken questions and assertions, and providing intelligible answers using synthesized speech. This system is also responsible for detecting when the user is addressing the system (versus chatting with a workmate) by verifying two types of events coming from the users:

\begin{itemize}
\item Using the keyword ``Celia" 
\item Looking directly at the camera
\end{itemize}

Once the user gets the system's attention using one of the above methods, he/she has 2 seconds to start phrasing his/her request. If nothing is heard within this interval, the system times out and resumes waiting for a new attention trigger.

\section{Conclusion and Future Work}

In this paper we detailed our preliminary work on keeping track of objects and people, and informing the user of various relations between them upon request. Our system is able to resolve the \textit{belonging} and allows the user ask about his/her/others' objects. While the system is currently part of a room, we believe a mobile robot has fewer privacy concerns than a static camera mounted on the ceiling. It can also look at objects from different points of view, thus reducing possible blind spots and occlusions. 

We are also planning to enable a proactive behavior where the robot follows people and watches which objects they are using. This way the robot will have a broader knowledge base of activities, and be able to answer questions about locations that are not in its current field of view. 

\bibliographystyle{aaai}
\bibliography{biblio}
 
\end{document}